\documentclass[runningheads]{llncs}
\usepackage{marvosym}
\usepackage{academicons,xcolor}
\usepackage{orcidlink}

\usepackage{caption}
\usepackage{graphicx,subfigure}
\usepackage{amsfonts,amssymb,amsmath}
\usepackage{graphicx}
\usepackage{multirow}
\usepackage{enumitem}

\usepackage[capitalize]{cleveref}
\usepackage{hyperref}

\begin{document}

\title{Deep Structure and Attention Aware Subspace Clustering}
\titlerunning{Deep Structure and Attention Aware Subspace Clustering}
%
\author{Wenhao Wu\orcidlink{0009-0008-0359-4082} \and
Weiwei Wang\textsuperscript{(\Letter)}\orcidlink{0000-0002-6985-2784} \and
Shengjiang Kong\orcidlink{0000-0002-8790-1455}}
\authorrunning{W. Wu et al.}
%
\institute{School of Mathematics and Statistics, Xidian University, Xi’an 710071, China \\
\email{wwwang@mail.xidian.edu.cn}}
\maketitle              
\begin{abstract}
Clustering is a fundamental unsupervised representation learning task with wide application in computer vision and pattern recognition. Deep clustering utilizes deep neural networks to learn latent representation, which is suitable for clustering. However, previous deep clustering methods, especially image clustering, focus on the features of the data itself and ignore the relationship between the data, which is crucial for clustering. In this paper, we propose a novel Deep Structure and Attention aware Subspace Clustering (DSASC), which simultaneously considers data content and structure information. We use a vision transformer to extract features, and the extracted features are divided into two parts, structure features, and content features. The two features are used to learn a more efficient subspace structure for spectral clustering. Extensive experimental results demonstrate that our method significantly outperforms state-of-the-art methods. Our code will be available at \url{https://github.com/cs-whh/DSASC}

\end{abstract}
\section{Introduction}

Clustering, aiming to partition a collection of data into distinct groups according to similarities, is one fundamental task in machine learning and has wide applications in computer vision, pattern recognition, and data mining. In recent years, a significant amount of research has been dedicated to Self-Expressive-based (SE) subspace clustering due to its effectiveness in dealing with high-dimensional data. SE-based subspace clustering primarily assumes that each data point can be expressed as a linear combination of other data points within the same subspace. It generally consists of two phases. In the first phase, a self-representation matrix is computed to capture similarity between data points. Subsequently, spectral clustering is employed to obtain the data segmentation. 

Learning an effective self-representation matrix is critical for clustering. Therefore, various regularization methods have been designed to pursue clustering-friendly properties of the self-representation matrix. For example, Sparse Subspace Clustering (SSC)~\cite{SSC} uses the $ \ell_1$ norm regularization to learn sparse representation. The Low-Rank Representation (LRR)~\cite{low_rank_sc}  uses the nuclear norm regularization to induce low-rank representation. You et al.~\cite{elastic} introduced an elastic net regularization to balance subspace preservation and connectivity. Although these methods have demonstrated impressive clustering performance, they are ineffective in dealing with complicated real data, which have non-linearity~\cite{DSC} and contain redundant information and corruptions.

Thanks to the powerful capability of Deep Neural Networks (DNNs) in representation learning, deep SE-based subspace clustering employs DNNs to integrate feature learning and self-representation, hoping that the latent features can reduce the redundancy and corruptions meanwhile satisfy the self-expressive assumption. Convolutional AutoEncoder (CAE) has been widely used for image clustering. For instance, the Deep Subspace Clustering network (DSC)~\cite{DSC} adds a self-representation layer between the convolutional encoder and the decoder to learn the self-representation matrix in the latent subspaces. However, the convolutional operations focus on local spatial neighborhoods, having limited ability to capture long-range dependencies in images.  To exploit the long-term dependence and extract more effective features, self-attention mechanisms and transformer architectures have emerged as leading techniques, delivering state-of-the-art performance in various computer vision tasks, including image classification~\cite{vit}, object recognition~\cite{DERT}, and semantic segmentation~\cite{strudel2021segmenter}. 

The traditional DNNs learn the feature representation of each data independently, ignoring the relationship between data points. In contrast, the Graph Convolutional Network (GCN)~\cite{GCN} learns the feature representation of connected data points collaboratively, which is helpful in exploiting the inherent feature of data. Based on GCN, the Attributed Graph Clustering (AGC)~\cite{AGC} and the Structural Deep Clustering Network (SDCN)~\cite{SDCN} have been proposed for clustering data. However, these works focus on data that naturally exist in graphs, such as citation and community networks, and few pay attention to image datasets.

The self-attention mechanisms and the transformer architectures are effective in capturing long-range dependencies in images. At the same time, the GCN is advantageous in exploring the inherent feature of similar data. That inspires us to propose the Deep Structure and Attention aware Subspace Clustering (DSASC) for image datasets. Specifically, the proposed network couples the Vision Transformer (ViT)~\cite{vit} and the GCN. The ViT is pre-trained on a large-scale image dataset using the state-of-the-art representation learning framework DINO (self-DIstillation with NO labels) ~\cite{DINO}, which is trained with two ViT of the same architecture, one called the teacher network and the other called the student network. The teacher network receives input from a global perspective, and the student network receives input from a local perspective. DINO expects the output of the student network and the teacher network to be consistent, thus learning the semantic features of the data. Once the training is completed, the teacher network is used for feature extraction. However, DINO only considers the information within each image, ignoring the cluster structure within similar images. To capture the cluster structure within similar images, we construct a K-Nearest Neighbor (KNN)~\cite{KNN} graph for intermediate features of ViT and use the GCN to extract the features from the graph perspective. We refer to the output of GCN as structural features and the output of ViT as content features.

To explore the underlying subspace structure of images, we learn the self-representation matrix for the content and structure features separately. It should be noted that the self-representation matrices of the content features and the structure features are learned collaboratively to facilitate each other. Finally, we use the fused self-representation matrix for spectral clustering. Our main contribution can be summarized as follow:
\begin{itemize}[label=$\bullet$]
    \item We propose a novel Deep Structure and Attention aware Subspace Clustering method that takes into account both data content and structural information, enhancing the performance of image clustering.
    \item We combining the strengths of a vision transformer and graph convolutional network  to enhance feature learning efficiency and clustering performance.
    \item Extended experiments show that the proposed method outperforms some state-of-the-art baselines.
\end{itemize}

\section{Proposed Method}
\label{sec:proposed}

\begin{figure}[t]
\includegraphics[width=\textwidth]{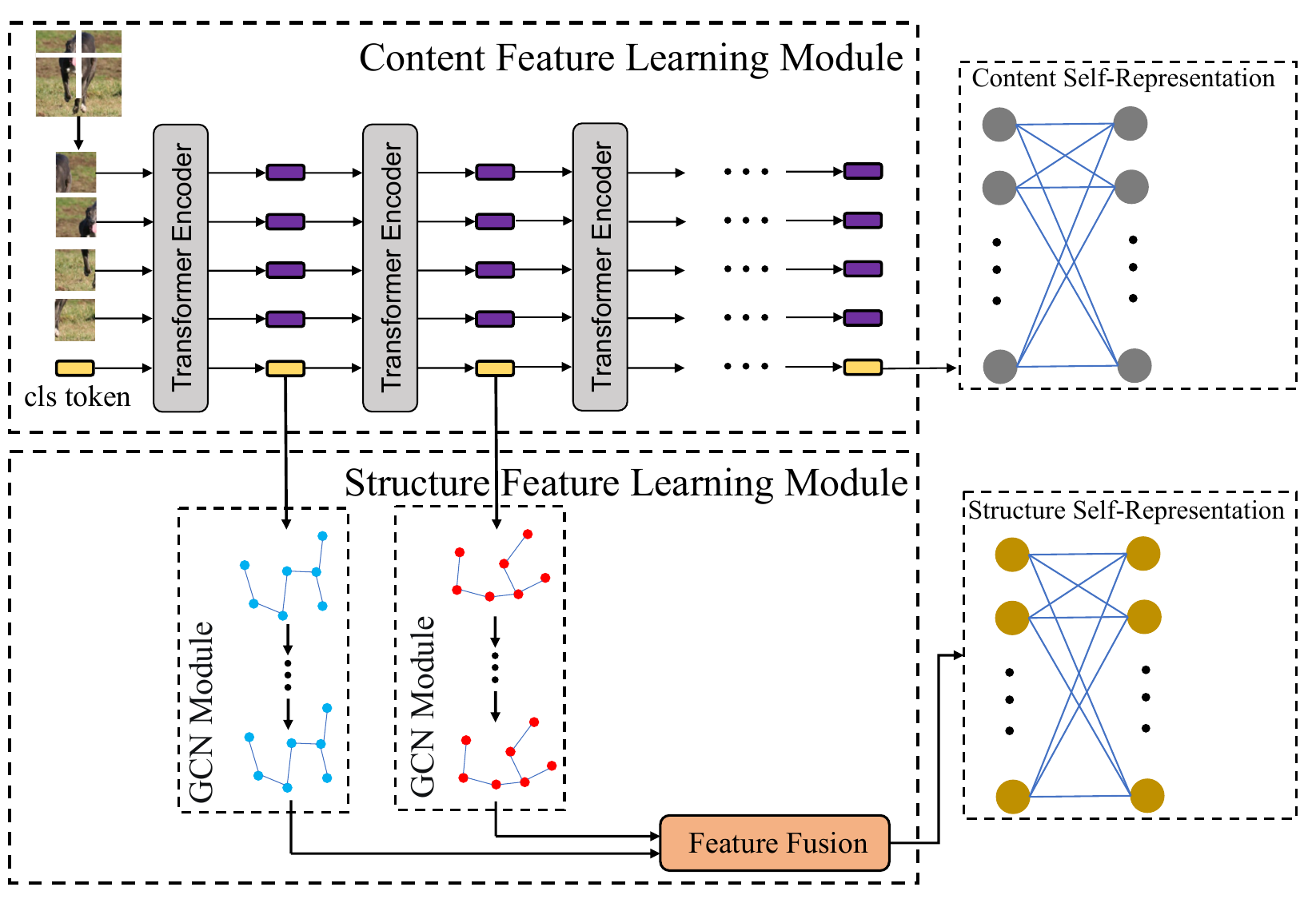}
\caption{The overall framework of our model.}\label{fig:network}
\end{figure}

ViT effectively captures long-range dependencies within images through self-attention mechanisms and the transformer architecture. We propose to use the pre-trained ViT to learn the content features of images. GCN can capture the structural features of data that exist in graphs form. To capture the cluster structure within similar images,  we organize the image as a graph using the KNN and use GCN to learn the structural features. Note that the ViT and the GCN are coupled. The ViT, comprising 12 layers, extracts content features across different depths. Our approach employs features from the third, sixth, and ninth ViT layers to create KNN graphs and utilizes GCN to learn structural features. Finally, we fuse the features extracted from multiple layers into a unified structural feature. The self-representation matrices for the content features and structure features are learned to explore the underlying subspace structure of images. Fig.~\ref{fig:network} illustrates our proposed network, which contains two coupled feature learning modules and two self-representation modules. 

The Content Feature Learning Module includes a ViT with an added learnable class embedding (cls token) and image patches, which are processed by the Transformer Encoder. The cls token's output serves as features (yellow patches). The Structure Feature Learning Module utilizes ViT's intermediate features and employs GCN for feature extraction. The Self-Representation Module is a fully-connected layer without bias and activation function to simulate the self-expressive process.

\begin{figure*}[t]
  \centering
  \subfigure[]{\label{fig:ssc_sim}\includegraphics[width=0.45\textwidth]{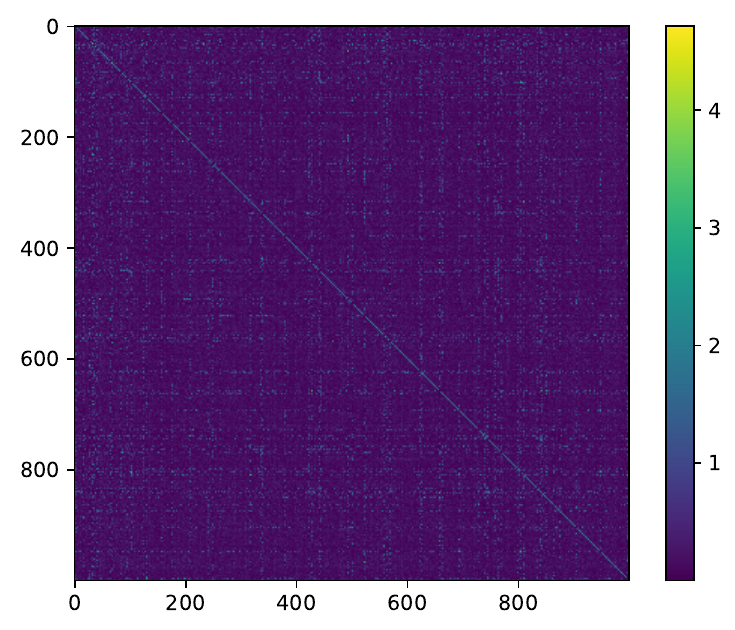}}
  \subfigure[]{\label{fig:fusion_self-expression}\includegraphics[width=0.48\textwidth]{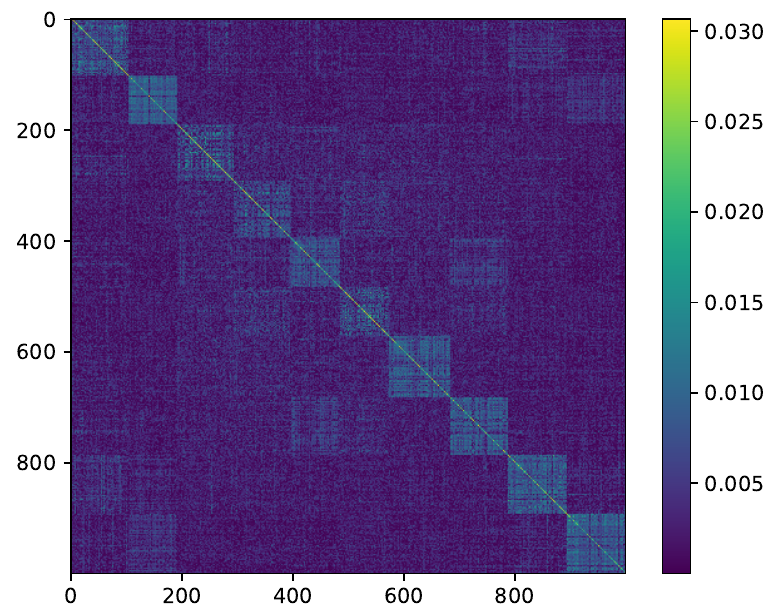}}
  \caption{(a) Affinity matrix obtained by SSC (b) Affinity matrix obtained by our method.}
  \label{fig:blockshow}
\end{figure*}

To show the effectiveness of our method, we select 1000 sample images for 10 clusters (100 samples for each cluster) from the dataset STL-10 and obtain the affinity matrices using SSC~\cite{SSC} and our method, respectively. Fig.~\ref{fig:blockshow} illustrates the results. Note that SSC computes the self-representation of the raw images directly, while our method learns the self-representation of the features. The affinity matrix is obtained by adding the absolute value of the self-representation and its transpose. SSC's affinity matrix lacks favorable diagonal blocks for subspace clustering. Conversely, our method's matrix displays distinct diagonal block structure.

\subsection{Content Feature Learning}
\label{sec:content}

DINO uses two networks with the same architecture during training, the teacher network $g_{\theta_t}$ and the student network $g_{\theta_s}$ with parameters ${\theta_t}$ and ${\theta_s}$, respectively. The output of both networks is followed by a softmax layer with temperature parameters to map the features into a probability distribution $P_t(x)$ and $P_s(x)$.

To train the two networks, DINO employs a multi-crop strategy~\cite{mulit-crop} to segment each image $x$ into multiple views, each with distinct resolutions. More precisely, starting from an input image $x$, a set $V$ of diverse views is generated, comprising two global views denoted as $x_1^g$ and $x_2^g$, along with several local views of lower resolutions. The student network mainly receives input from the local view, while the teacher network receives input from the global view.

At each iteration, the weights $\theta_s$ are updated through a gradient descent to minimize the cumulative cross-entropy loss over a minibatch of input images.
\begin{equation}
    \min\limits_{\theta_s} \sum\limits_{x \in \{x_1^g, x_2^g\}} \sum\limits_{\substack{x^\prime \in V \\ x^\prime \neq x}} -P_t(x) \log P_s(x^\prime)
\end{equation}

The parameters $\theta_t$ of the teacher network are derived by using the exponential moving average (EMA) technique on the student parameters $\theta_s$.

\begin{equation}
    \theta_t \leftarrow \lambda \theta_t + (1-\lambda) \theta_s
\end{equation}
where the value of $\lambda$ is selected based on a cosine schedule. To avoid training collapse, the centering operation is conducted to the teacher network.
\begin{equation}
g_t(x) \leftarrow g_t(x)+c
\end{equation}
where $c$ is updated every mini-batch.
\begin{equation}
c \leftarrow m c+(1-m) \frac{1}{|B|} \sum_{i: x_i \in B} g_{\theta_t}\left(x_i\right)
\end{equation}
where $m > 0$ is a rate parameter and $B$ is the batch size. In the training process,  the local view is required to match the global view, thus compelling the network to learn semantic features rather than low-level features. After DINO training, the $g_{\theta_t}$ is used for feature extraction, and the content features are obtained using the formula $F_A = g_{\theta_t}(x)$.


\subsection{Structure Feature Learning}
\label{sec:graph}

\subsubsection{ViT Intermediate Feature Extraction.}

ViT consists of a sequence of Transformer Encoders that take a sequence as input and produce an output sequence of equal length.
The output sequence has been feature extracted using the attention mechanism. Let the raw image be $X$, the image is divided into $b$ blocks, and an additional cls token has been added. Consider these $b+1$ blocks as the input sequence, denoted as $M_0 \in \mathbb{R}^{(b+1) \times d}$, where $d$ is the dimension of each block. Feeding data into a ViT can be formalized as follows.
\begin{equation}
    M_{i+1} = T(M_i), i=0,1,...,L_1
\end{equation}
where $T$ is the transformer encoder,  $L_1$ is the number of layers of ViT. We extracted some intermediate layers and let the indices of the selected layers be $S$. Then the extracted features can be denoted as $\{M_k[0,:], k \in S\}$, where $M_k[0,:]$ refers to the first row of the matrix $M_k$, which is the feature corresponding to the cls token. We denote the features as $H_1, H_2,..., H_t $ with $t = |S|$ for simplicity, and construct KNN graphs measuring by cosine similarity for each feature and obtain adjacency matrix: $A_1, A_2,..., A_t$.

\subsubsection{GCN Module.}

To effectively extract features and learn node representations from graph structures, we utilize the GCN module for feature extraction. The GCN module can be expressed as follows.
\begin{equation}
\begin{aligned}
  &  H_i^{(0)} = H_i, \quad i=1,2...,t\\
H_i^{(l)} = \phi (D_i^{-\frac{1}{2}} \Tilde{A}_i & D_i^{-\frac{1}{2}} H_i^{(l-1)}W_i^{(l-1)}), \quad l=1,2,...,L_2
\end{aligned}
\end{equation}
where $H_i$ is the feature obtained from the ViT intermediate layer. $W_i^{(l-1)}$ is the learnable parameter, $D_i$ is the degree matrix, $L_2$ is the number of layers of graph convolution, $\Tilde{A}_i = A_i + I$ takes into account the self-connectivity of nodes and $\phi$ is the activation function.

\subsubsection{Features Fusion.}
To fuse the learned structure features, we use an adaptive method with learnable parameters $W_i$. This method avoids some deviations caused by artificial design, and allows the network to identify which feature is more important.
\begin{equation}
    F_S = \sum\limits_{i=1}^t H_i^{(L_2)} W_i
\end{equation}

\subsection{Training strategy}
\label{sec:training strategy}

To explore the internal subspace structure of the data, we use a fully-connected layer without bias and activation function to simulate the self-expressive process and learn the self-expressive matrix of content features. In particular, we minimize the following objective.
\begin{equation}
    \label{eq:content_loss}
    \min _{C_A}\left\|C_A\right\|_1+\lambda_1\left\|F_A-F_A C_A\right\|_F^2
\end{equation}
where $ C_A \in \mathbb{R}^{n \times n}$ is the self-expreesive matrix of content. Similar to the content self-representation, we adopt the same way to learn the self-representation matrix $C_S$ of structural features.
\begin{equation}
    \min _{C_S}\left\|C_S\right\|_1+\lambda_2\left\|F_S-F_S C_S\right\|_F^2
\end{equation}

After acquiring the $C_A$ and $C_S$, we combine them to construct the final affinity graph $C_F$ for spectral clustering. Instead of adopting the method proposed by previous work~\cite{AASSC}, we adopt a simple yet effective fusion strategy: we add two self-expressive matrix to obtain the fused self-expressive matrix. 
\begin{equation}
    C_F = C_A + C_S
\end{equation}
This approach can enhance self-expressive consistency and alleviate the error caused by random noise. We also add a sparsity constraint to $C_F$, ensuring that a meaningful representation is learned.
\begin{equation}
    \min_{C_F} \left\| C_F \right\|_1
\end{equation}

The total loss function is:
\begin{equation}
\label{eq:total_loss}
\begin{aligned}
    \mathcal{L} = \left\|C_A\right\|_1 & +  \lambda_1\left\|F_A-F_A C_A\right\|_F^2 + \left\|C_S\right\|_1 \\
    & +\lambda_2\left\|F_S-F_S C_S\right\|_F^2 + \left\| C_F \right\|_1
\end{aligned}
\end{equation}

\section{Experiments}
\label{sec:exp}
In this section, we first present the datasets used in our experiment, the implementation details, and the baselines. Then we present the experimental results and conduct an analysis to validate the efficacy of our proposed method.

\subsection{Experiments Settings}
\subsubsection{Datasets.} We evaluate our method on four image benchmark datasets including CIFAR10, STL-10, Fashion-MNIST and CIFAR100. The CIFAR-10 dataset comprises 60,000 RGB images depicting 10 distinct objects. The STL-10 dataset consists of 13,000 RGB images, depicting 10 distinct objects, with each image having a resolution of 96×96 pixels. The Fashion-MNIST dataset consists of 70,000 grayscale images of various fashion products, classified into 10 categories, with a resolution of 28×28 pixels. The CIFAR100 dataset is a more challenging dataset that consists of 60,000 images for 100 different objects. There are 600 images per class. 

For the computation efficiency of the self-representation matrix, we randomly take 1000 images from CIFAR10, STL-10, and Fashion-MNIST for cluster analysis. For the CIFAR100 dataset, we randomly take 3000 images of the 100 objects (300 images for each object) for clustering.

\begin{table}[t]
  \centering
  \caption{Clustering results of different methods on benchmark datasets. The top-performing results are highlighted in bold, while the second-best results are marked with an underline.}
    \begin{tabular}{ccccccccc}
    \hline
    \multirow{2}[2]{*}{Method} & \multicolumn{2}{c}{CIFAR10} & \multicolumn{2}{c}{STL10} & \multicolumn{2}{c}{Fashion-MNIST} & \multicolumn{2}{c}{CIFAR100} \\
        & ACC & NMI & ACC & NMI & ACC & NMI & ACC & NMI \\
    \hline
    k-means & 0.2290  & 0.0840  & 0.1920  & 0.1250  & 0.4740 & 0.5120  & 0.1300  & 0.0840  \\
    LSSC & 0.2114  & 0.1089  & 0.1875  & 0.1168  & 0.4740  & 0.5120  & 0.1460  & 0.0792  \\
    LPMF & 0.1910  & 0.0810  & 0.1800  & 0.0960  & 0.4340  & 0.4250  & 0.1180  & 0.0790  \\
    DEC & 0.3010  & 0.2570  & 0.3590  & 0.2760  & 0.5180  & 0.5463  & 0.1850  & 0.1360  \\
    IDEC & 0.3699  & 0.3253  & 0.3253  & 0.1885  & 0.5290  & 0.5570  & 0.1916  & 0.1458  \\
    DCN & 0.3047  & 0.2458  & 0.3384  & 0.2412  & 0.5122  & 0.5547  & 0.2017  & 0.1254  \\
    DKM & 0.3526  & 0.2612  & 0.3261  & 0.2912  & 0.5131  & 0.5557  & 0.1814  & 0.1230  \\
    VaDE & 0.2910  & 0.2450  & 0.2810  & 0.2000  & 0.5039  & 0.5963  & 0.1520  & 0.1008  \\
    DSL & 0.8340  & 0.7132  & 0.9602 & 0.9190 & 0.6290  & 0.6358  & 0.5030  & 0.4980  \\
    EDESC & 0.6270 & 0.4640 & 0.7450  & 0.6870 & 0.6310 & 0.6700 & 0.3850  & 0.3700  \\
    CDEC & 0.5640  & 0.5778  & 0.7328  & 0.7183  & 0.5351  & 0.5404  & -   & - \\
    DML & \underline{0.8415}  & \underline{0.7170}  & \underline{0.9645} & \underline{0.9211} & 0.6320  & 0.6480  & \underline{0.5068}  & \underline{0.5019}  \\
    SENet & 0.7650 & 0.6550  & 0.8232  & 0.7541  & \textbf{0.6970} & \underline{0.6630} & -   & - \\
    Our & \textbf{0.8740 } & \textbf{0.8059 } & \textbf{0.9740 } & \textbf{0.9491 } & \underline{0.6680}  & \textbf{0.6739 } & \textbf{0.5200 } & \textbf{0.6865 } \\
    \hline
    \end{tabular}%
  \label{tab:results}%
\end{table}%

\subsubsection{Implementation details.}
We use the ViT-S/8 architecture, which was trained on ImageNet-1k (without labels) for content feature extraction. We use the pre-trained parameters provided by the authors. During our training, the parameters of the ViT are frozen. The GCN module consists of 2 layers of graph convolution, and the ReLU activation function is used. To construct the KNN graph, we set $K=10$. The $\lambda_1$ and $\lambda_2$ in Eq.~\ref{eq:total_loss} are set at $1$. The Adam optimizer is used to minimize the loss, and the learning rate is set to $10^{-5}$. We train for 2000 epochs and report the final results.

\subsubsection{Evaluation Metrics.}
We use two commonly used metrics to evaluate clustering performance, including clustering accuracy (ACC) and normalized mutual information (NMI).
The clustering performance is positively correlated with higher values obtained from these metrics. Specifically, ACC is calculated by
\begin{equation}
A C C=\max _m \frac{\sum_{i=1}^n \mathbf{1}\left\{l_i=m\left(c_i\right)\right\}}{n}
\end{equation}
where $l_i$ denotes the ground truth, $c_i$ is the predicted assignment. $\mathbf{1}$ is the
indicator function. And the $m$ establishes a mapping between the ground-truth labels and predicted assignments, encompassing all possible one-to-one correspondences. Notably, the Hungarian algorithm~\cite{kuhn1955hungarian} is a highly effective computational method for efficiently determining the optimal mapping function. The NMI is calculated by
\begin{equation}
N M I=\frac{I(\boldsymbol{l} ; \boldsymbol{c})}{\max \{H(\boldsymbol{l}), H(\boldsymbol{c})\}}
\end{equation}
where the $ \boldsymbol{l} = (l_1, l_2, ... , l_n)$ is the ground-truth label of $n$ samples and $\boldsymbol{c} = (c_1, c_2,...,c_n)$  is the predicted assignment. $H(\cdot)$ and $I(\cdot)$ is entropy and mutual information, respectively.

\subsubsection{Comparison algorithms.}
We compare our method DSASC with three traditional methods: k-means~\cite{kmeans}, large-scale spectral clustering (LSSC)~\cite{LSSC}, locality preserving non-negative matrix factorization (LPMF)~\cite{LPMF}, and ten deep clustering methods: deep embedding clustering (DEC)~\cite{DEC}, improved deep embedding clustering (IDEC)~\cite{IDEC},   deep clustering network (DCN)~\cite{towardskmeans}, deep k-means (DKM)~\cite{deepkmeans}, variational deep embedding (VaDE)~\cite{VaDE}, deep successive subspace learning (DSL)~\cite{DSL},  efficient deep embedded subspace clustering (EDESC)~\cite{cai2022efficient}, contrastive deep embedded clustering (CDEC)~\cite{CDEC}, deep multi-representation learning (DML)~\cite{DML}, self-expressive network (SENet)~\cite{SENet}.

\begin{figure*}[t]
  \centering
  \subfigure[STL10 dataset]{\label{fig:stl10_vision}\includegraphics[width=\textwidth]{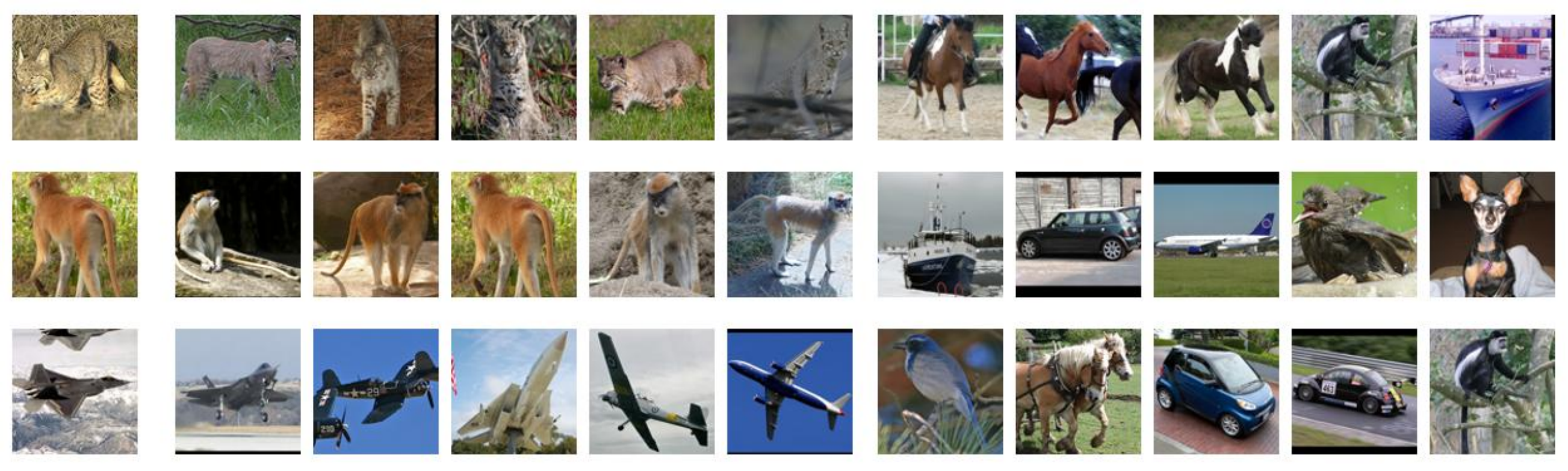}}
  \subfigure[Fashion-MNIST dataset]{\label{fig:fmnist_vision}\includegraphics[width=\textwidth]{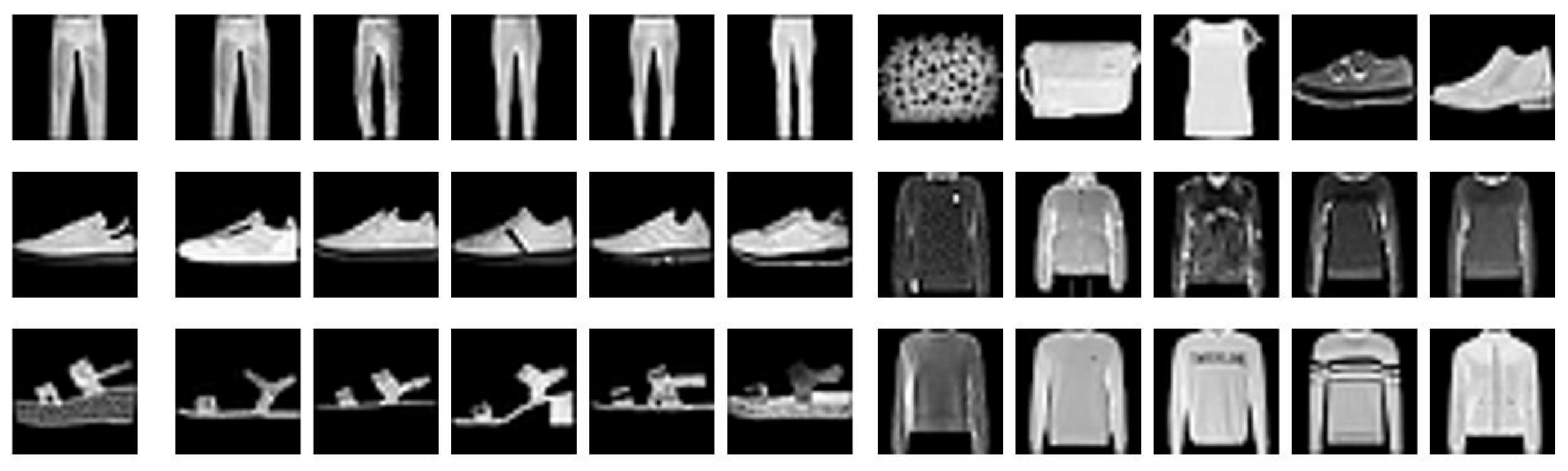}}
  \caption{The first column is the anchor image. We sort the numerical values in the self-expressive matrix. Where columns 2 to 6 are images corresponding to top-5 values (most similar) and columns 7 to 11 are images corresponding to bottom-K values (least similar).}
  \label{fig:vision}
\end{figure*}

\subsection{Results and Discussion}

We use two commonly used metrics to evaluate clustering performance, including clustering accuracy (ACC) and normalized mutual information (NMI). The clustering results are presented in Table~\ref{tab:results}, where the best results are emphasized in bold the second best results are marked with an underline. As evident from the table, DSASC significantly improves the clustering performance. DSASC improves ACC and NMI of DML in CIFAR10, respectively, by 3.25\% and 8.89\%. The non-significant improvement in clustering performance of DSASC on Fashion-MNIST is due to the fact that Fashion-MNIST is gray-scale images, while our ViT is pre-trained on color images and suffers from domain shift. But our method still achieves competitive results on Fashion-MNIST. That suggests that combining structure features and content features effectively improve the discrimination of the feature representation and clustering performance.

\begin{figure*}[t]
  \centering
  \subfigure[]{\label{fig:cifar10_acc_nmi_with_epoch}\includegraphics[width=0.24\textwidth]{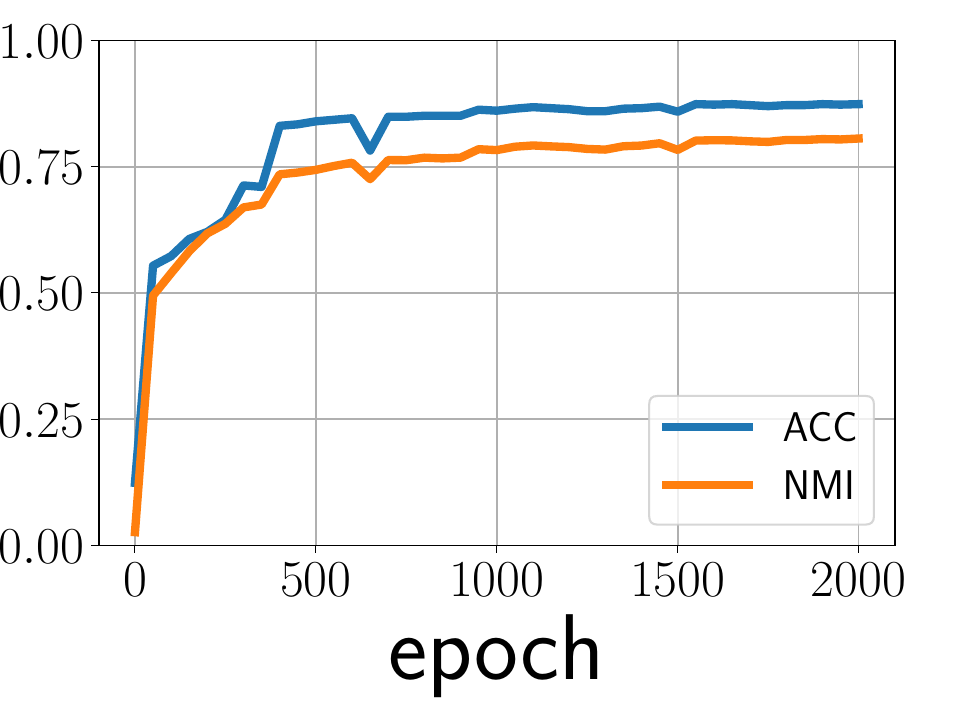}}
  \subfigure[]{\label{fig:cifar10_k}\includegraphics[width=0.24\textwidth]{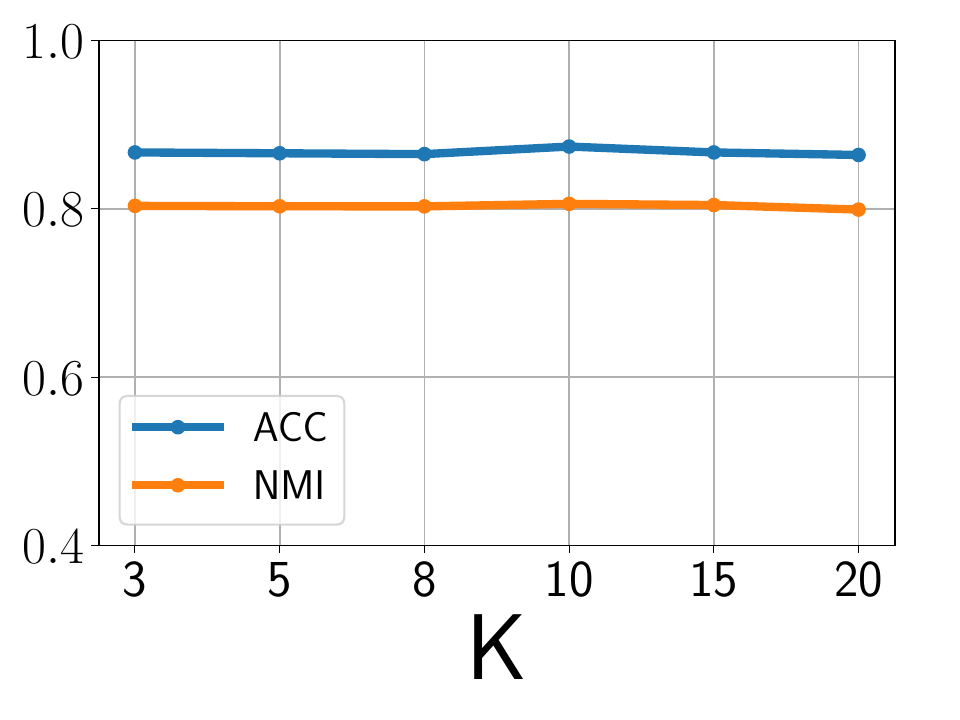}}
  \subfigure[]{\label{fig:cifar10lambda1}\includegraphics[width=0.24\textwidth]{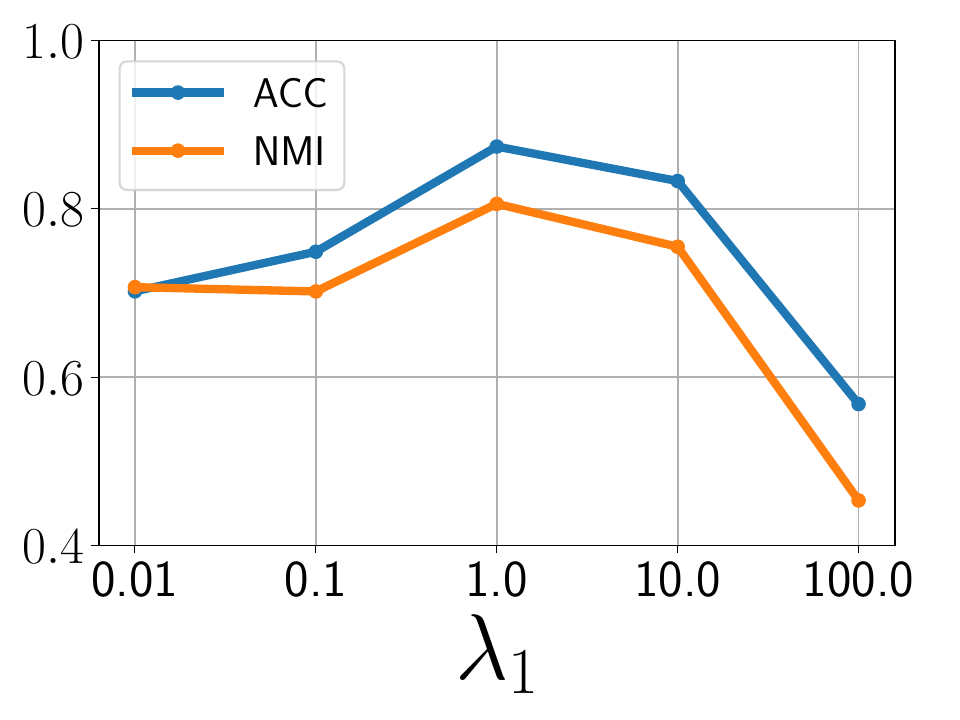}}
  \subfigure[]{\label{fig:cifar10lambda2}\includegraphics[width=0.24\textwidth]{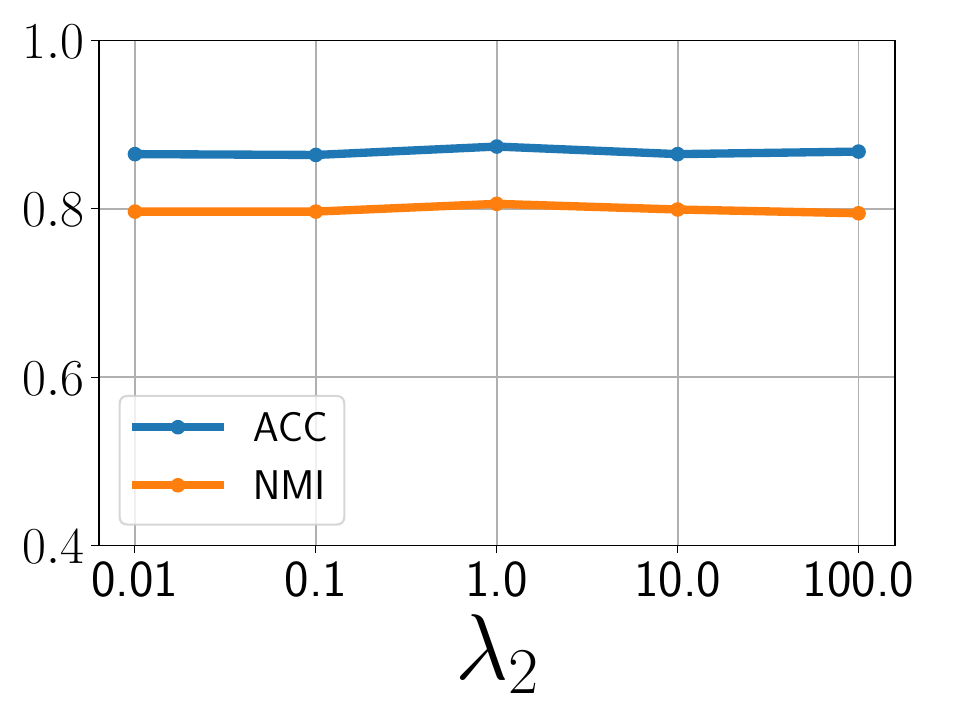}}
  \subfigure[]{\label{fig:fmnist_acc_nmi_with_epoch}\includegraphics[width=0.24\textwidth]{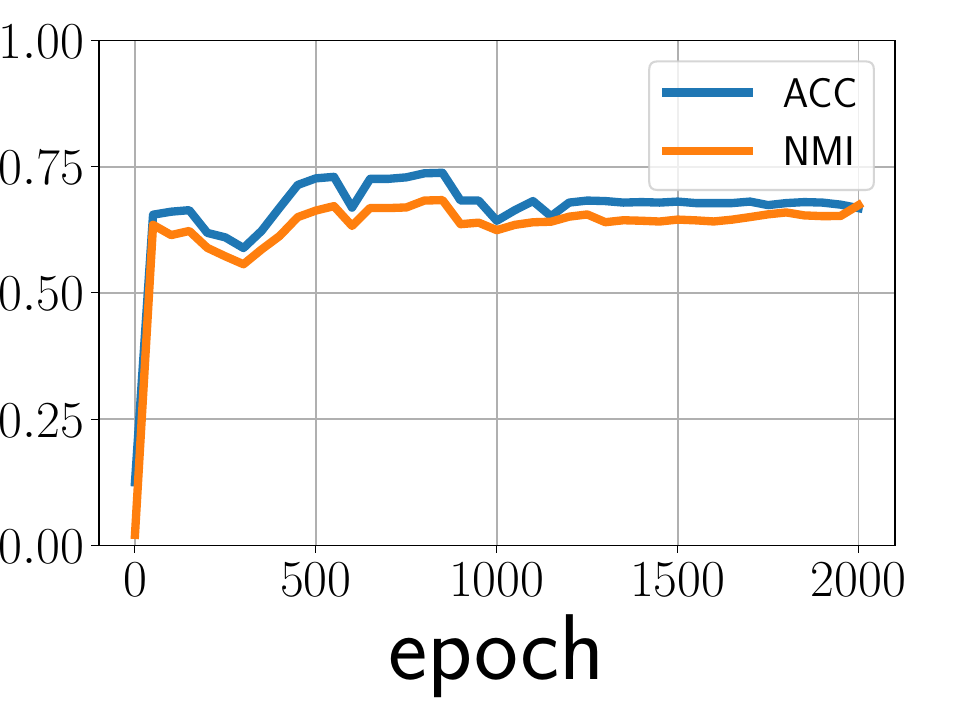}}
  \subfigure[]{\label{fig:fmnist_K}\includegraphics[width=0.24\textwidth]{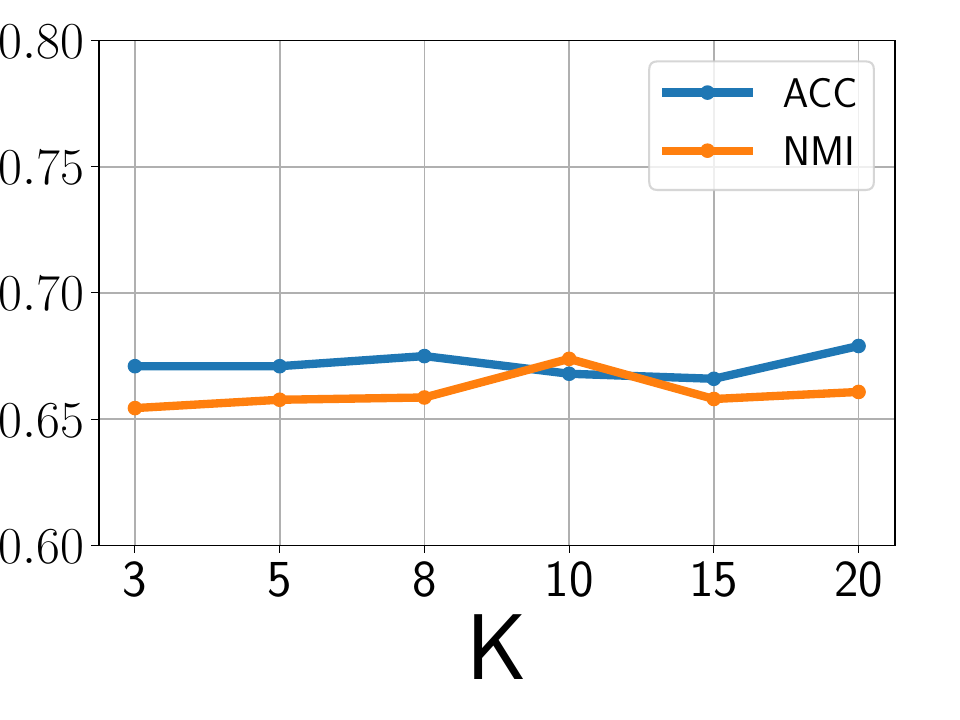}}
  \subfigure[]{\label{fig:fmnist_lambda1}\includegraphics[width=0.24\textwidth]{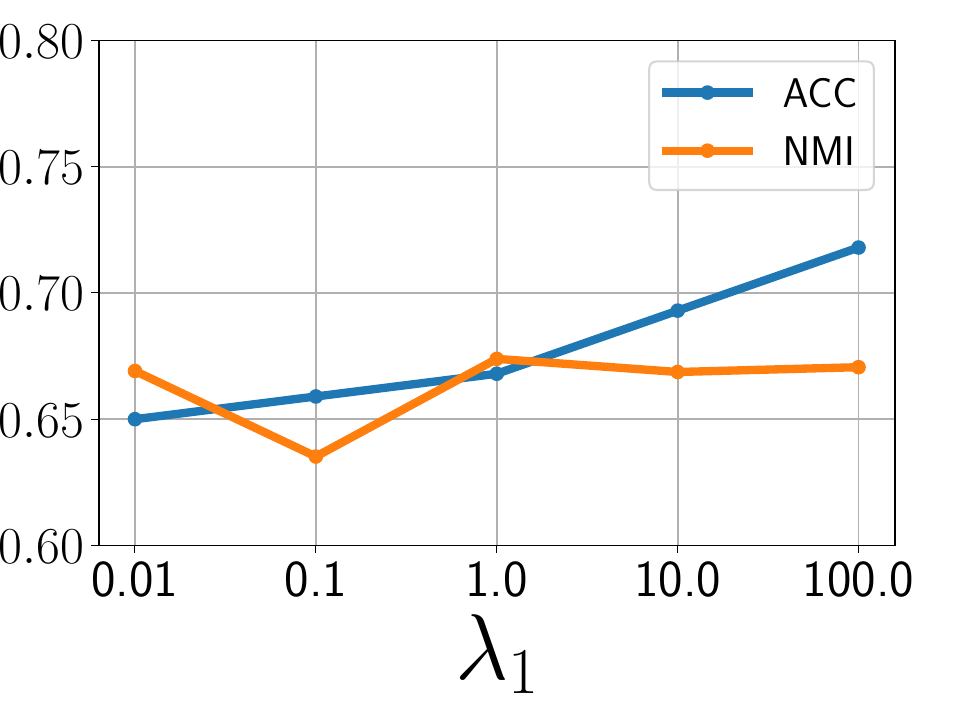}}
  \subfigure[]{\label{fig:fmnist_lambda2}\includegraphics[width=0.24\textwidth]{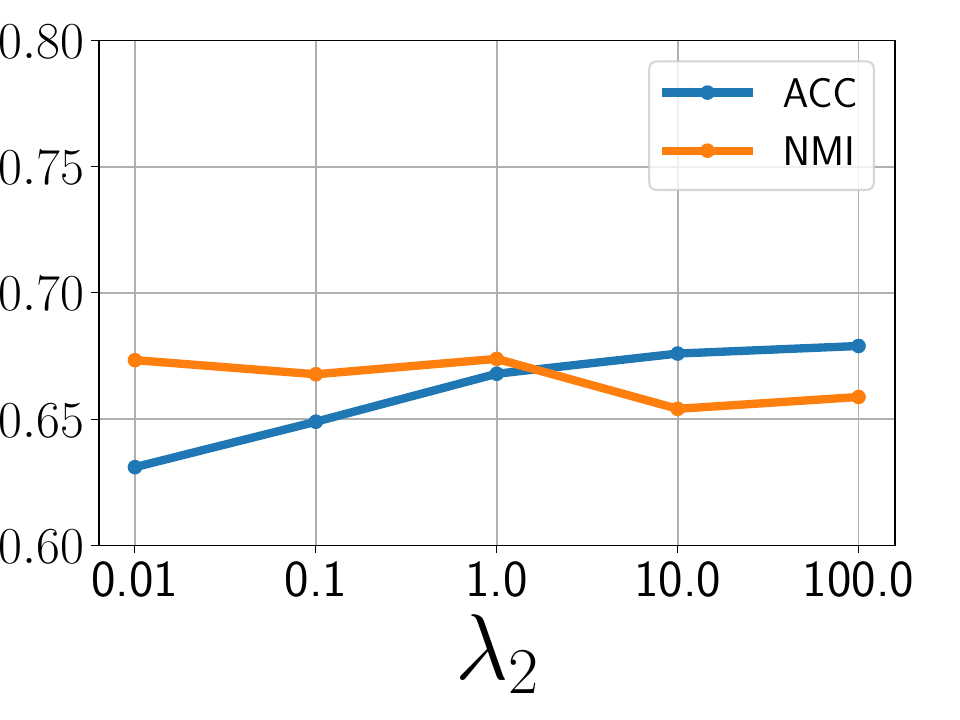}}
  \caption{Cluster convergence analysis and parameter sensitivity analysis. The first row shows the results for CIFAR10 and the second row shows the results for Fashion-MNIST. (a)(e) Changes of clustering ACC and NMI with training epochs. (b)(f) The influence of different K on the clustering results. (c)(g) The influence of different $\lambda_1$ on the clustering results. (d)(h) The influence of different $\lambda_2$ on the clustering results.}
  \label{fig:curve}
\end{figure*}

\begin{table}[t]
  \centering
  \caption{The ablation experiments results. The top-performing results are highlighted in bold.}
    \begin{tabular}{ccccc}
    \hline
       \multirow{2}[2]{*}{Datasets} & \multicolumn{2}{c}{Content SE} & \multicolumn{2}{c}{Our Method} \\
        & ACC & NMI & ACC & NMI \\
    \hline
    CIFAR10 & 0.8370  & 0.7430  & \textbf{0.8740}  & \textbf{0.8059}  \\
    STL10 & 0.9380  & 0.8886  & \textbf{0.9740}  & \textbf{0.9491}  \\
    Fashion-MNIST & 0.6560  & 0.6741  & \textbf{0.6680}  & \textbf{0.6739}  \\
    CIFAR100 & 0.3527  & 0.5220  & \textbf{0.5200}  & \textbf{0.6865} \\
    \hline
    \end{tabular}%
  \label{tab:ablation}%
\end{table}%

In order to confirm the efficacy of our proposed method, we conduct ablation experiments on structure feature learning module. We remove the structural feature learning module and learn the content self-representation matrix $C_A$ using equation~\ref{eq:content_loss}, then perform spectral clustering. We denote the method as Content SE. Table~\ref{tab:ablation} presents the results. It can be seen that joint content features and structural features can significantly improve the clustering performance.

To further examine the quality of the self-expressive matrix obtained through our approach, we randomly selected three anchor images from the STL10 and Fashion-MNIST datasets, respectively. Subsequently, we have determined the five largest and five smallest representation coefficients of anchor images derived from the self-representation matrix $C_F$. These coefficients indicate the most similar and least similar images, respectively. This analysis provides insights into the effectiveness of the matrix and its ability to capture meaningful patterns. The result is shown in Fig.~\ref{fig:vision}. Where the first column is the anchor images, the next five are the most similar images, and the last are the most dissimilar images. It can be seen that our method can effectively learn the subspace structure, leading to satisfactory clustering outcomes.

Fig.~\ref{fig:cifar10_acc_nmi_with_epoch} and Fig.~\ref{fig:fmnist_acc_nmi_with_epoch} show the change of ACC and NMI during training, and it can be seen that the accuracy gradually improves as the training proceeds, which indicates that our method can effectively converge.

We additionally investigate the impact of hyperparameters on the experimental results. The hyperparameters include the value of $K$ in the GCN Module, $\lambda_1$, and $\lambda_2$ in Equation~\ref{eq:total_loss}. In the following experiments, the default value of $K$ is $10$, the default value of  $\lambda_1$ and $\lambda_2$ is $1$, and when one parameter is investigated, the other parameters are kept as default. We select $K \in \{3,5,8,10,15,20\}$ and  $\lambda_1,\lambda_2 \in\{0.01,0.1,1,10,100\} $ to evaluate their influence. The corresponding ACC and NMI of the clustering results are presented in \cref{fig:cifar10_k,fig:cifar10lambda1,fig:cifar10lambda2} for the dataset CIFAR10 and \cref{fig:fmnist_K,fig:fmnist_lambda1,fig:fmnist_lambda2} for the dataset Fashion-MNIST, respectively. As we can see from these figures, our method is not sensitive to the number of neighbors selection, and $\lambda_2$ also has less impact on the clustering performance. The content self-representation coefficient $\lambda_1$ dominates the clustering results. When $\lambda_1$ increases, the clustering performance of CIFAR10 decreases, while the trend is the opposite in Fashion-MNIST. We chose $\lambda_1=1$ for all datasets in our experiments, which produced competitive results.

\section{Conclusion}
\label{sec:con}

We propose a Deep Structure and Attention aware Subspace Clustering framework. Specifically, to discover the latent subspace structure of images, our network learns content features and structure features, as well as their sparse self-representation. Extended experiments show that our proposed method is highly effective in unsupervised representation learning and clustering.

\section*{Acknowledgements}
The authors would like to thank the editors and the anonymous reviewers for their constructive comments and suggestions. This paper is
supported by the National Natural Science Foundation of China (Grant
Nos. 61972264, 62072312), Natural Science Foundation
of Guangdong Province (Grant No. 2019A1515010894) and Natural
Science Foundation of Shenzhen (Grant No. 20200807165235002).

\bibliographystyle{splncs04}
\bibliography{samplepaper}
\end{document}